\pdfoutput=1

\documentclass[10pt,twocolumn,letterpaper]{article}
\newcommand{\textred}[1]{\textcolor{black}{#1}}
\newcommand{\tth}{\textsuperscript{th}}
\newcommand{\name}{\mbox{Wave-Former}}

\newcommand{\eqsize}{\small}

\newcommand{\cut}[1]{}
\usepackage[table]{xcolor}     
\usepackage{array}             
\usepackage{booktabs}         
\usepackage{multirow}         
\usepackage{makecell}         
\usepackage{caption}          
\usepackage{geometry}  
\usepackage{xr}
\usepackage[dvipsnames]{xcolor}

\usepackage{amsmath}

\DeclareMathOperator{\acos}{acos}

\usepackage{cvpr}              % To produce the CAMERA-READY version
\definecolor{cvprblue}{rgb}{0.21,0.49,0.74}
\usepackage[pagebackref,breaklinks,colorlinks,allcolors=cvprblue]{hyperref}

\def\paperID{10361} % *** Enter the Paper ID here
\def\confName{CVPR}
\def\confYear{2026}

\title{\vspace{-0.1in} \name: Through-Occlusion 3D Reconstruction \\ via Wireless Shape Completion \vspace{-0.1in}}

\author{ Laura Dodds$^1$, Maisy Lam$^1$, Waleed Akbar$^1$, Yibo Cheng$^1$, Fadel Adib$^{1,2}$\\
$^1$ - Massachusetts Institute of Technology, $^2$ - Cartesian Systems \\
{\tt\small \{ldodds, mllam, wakbar, yiboc, fadel\}@mit.edu}\\
}

\usepackage{enumitem}
\setlist[enumerate]{itemindent=12pt,
                    leftmargin=1pt, itemsep=1pt }
\setlist[itemize]{itemindent=12pt,
                    leftmargin=1pt, itemsep=1pt}
\begin{document}

\twocolumn[{%
\renewcommand\twocolumn[1][]{#1}%
\maketitle
\begin{center}
\vspace{-0.125in}
    \centering
    \captionsetup{type=figure}
    \includegraphics[width=\textwidth]{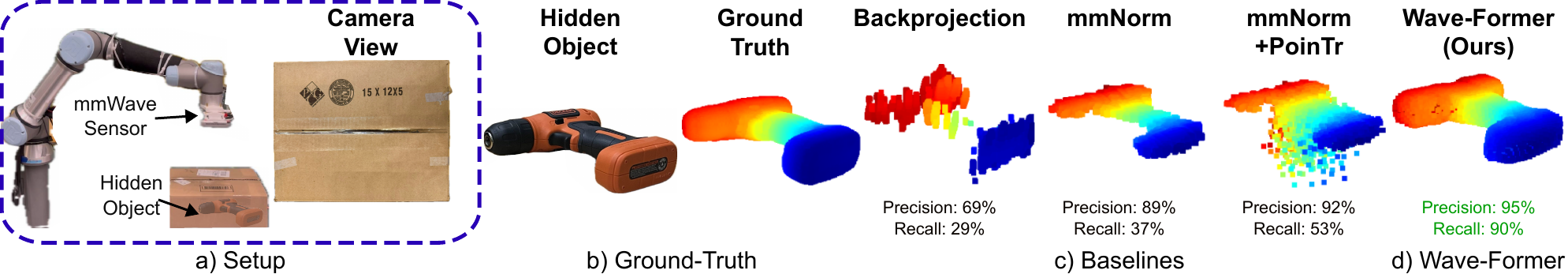}
    
    \vspace{-0.12in}
        \captionof{figure}{\footnotesize{\textbf{\name:} \footnotesize{a) \name\ can reconstruct objects hidden inside closed boxes using a millimeter-wave sensor. b) The ground-truth RGB and point cloud of a hidden object. c) Two state-of-the-art mmWave reconstruction methods (Backprojection and mmNorm) cannot reconstruct the entire object shape. Simply applying vision-based shape completion models (PoinTr) produces noisy reconstructions. d) By incorporating mmWave physical properties, \name\ can reconstruct the complete 3D shape of a hidden object from mmWave signals. }}}
    \label{fig:intro}
    
    \vspace{-0.075in}
\end{center}%
}]

\maketitle

\begin{abstract}

\vspace{-0.15in}
We present \name, a novel method capable of high-accuracy 3D shape reconstruction for completely occluded, diverse, everyday objects. 
This capability can open new applications spanning robotics, augmented reality, and logistics. Our approach leverages millimeter-wave (mmWave) wireless signals, which can penetrate common occlusions and reflect off hidden objects. In contrast to past mmWave reconstruction methods, which suffer from limited coverage and high noise, \name\ introduces a physics-aware shape completion model capable of inferring full 3D geometry.
At the heart of \name's design is a novel three-stage pipeline which bridges raw wireless signals with recent advancements in vision-based shape completion by incorporating physical properties of mmWave signals. The pipeline proposes candidate geometric surfaces, employs a transformer-based shape completion model designed specifically for mmWave signals, and finally performs entropy-guided surface selection. This enables \name\ to be trained using entirely synthetic point-clouds, while demonstrating impressive generalization to real-world data.
In head-to-head comparisons with state-of-the-art baselines,  \name\ raises recall from 54\% to 72\% while maintaining a high precision of 85\%.

\end{abstract}

\vspace{-0.2in}
\section{Introduction}
\vspace{-0.05in}

Reconstructing the 3D geometry of a fully occluded object, such as one inside a closed box or beneath clutter, is an open challenge in computer vision. This capability could enable numerous applications spanning robotic manipulation, augmented reality, and shipping and logistics. However, optical sensors, such as cameras and LiDARs, are inherently limited in such scenarios. In contrast, radio-frequency signals, such as millimeter-waves (mmWave), can traverse through occlusions and reflect off hidden objects. This ability has inspired recent interest in using these reflections for partial reconstruction of hidden objects~\cite{LauraEPFL, mmnorm}.

\begin{figure}[t]
\centering
    \vspace{0.05in}
    \includegraphics[width=0.85\linewidth]{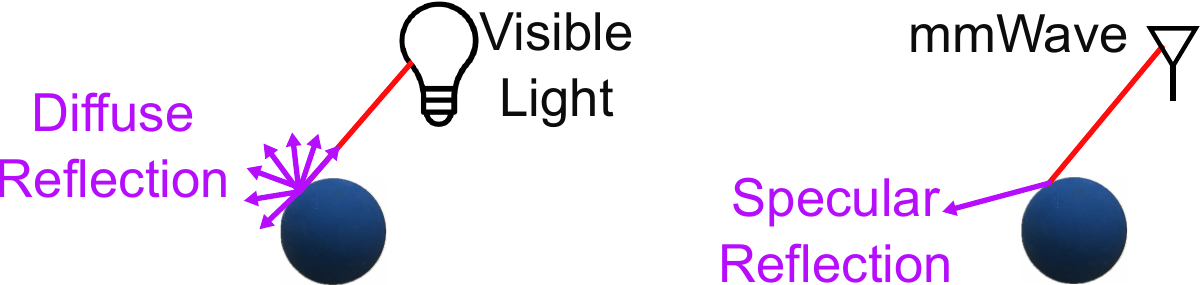} 
    \vspace{-0.15in}
    \caption{\footnotesize{\textbf{Specular reflection.} \textnormal{Unlike visible light, which primarily scatters, mmWave signals undergo primarily specular (mirror-like) reflections.}}}

    \label{fig:specular}
    \vspace{-0.2in}
    \end{figure}

However, estimating the complete 3D shape of hidden objects with mmWave signals remains challenging. This limitation arises from the unique physical properties of mmWave reflections. Unlike visible light, which scatters diffusely off most surfaces, mmWave signals reflect primarily in a specular (mirror-like) manner~\cite{SpecularReflection} (Fig.~\ref{fig:specular}). As a result, large portions of the surface will reflect signals away from the mmWave sensor, causing them to be effectively invisible and severely limiting the coverage. This issue is further compounded by the fact that mmWave signals are significantly noisier\footnote{Compared to the Kinect depth camera, point clouds produced with mmWave signals have roughly 5$\times$ more noise~\cite{mmnorm,kinect-acc}.} and capture surface reflections at much lower resolutions, making complete 3D reconstruction even more difficult. For example, we show two state-of-the-art mmWave reconstruction methods in Fig.~\ref{fig:intro} c, Backprojection~\cite{LauraEPFL} and mmNorm~\cite{mmnorm}, on a real-world, through-occlusion experiment. Both methods only capture a small portion of the object. 

To overcome this, one natural approach is to apply existing vision-based shape completion models~\cite{yu2021pointr, seedformer, snowflake} to these partial mmWave reconstructions. Unfortunately, this strategy fails to produce reliable reconstructions, since these existing models are designed for visible-light sensors with high coverage and resolution, and do not account for the unique physical properties of mmWave reflections. For instance, we show an example in Fig.~\ref{fig:intro}c where applying a state-of-the-art vision-based model, PoinTr~\cite{yu2021pointr}, to a mmWave partial point cloud produces a highly inaccurate and noisy reconstruction. 

In this paper, we present \name, the first complete mmWave 3D reconstruction method that can estimate the geometry of diverse objects through occlusions. \name\ bridges the gap between wireless sensing and modern shape completion techniques by embedding unique mmWave physics into the learning process. In Fig.~\ref{fig:intro}d, we demonstrate that \name\ improves reconstruction quality on a real-world through-occlusion example, where the object is completely hidden from view. 

Our core idea is a physics-aware training framework that embeds the physical characteristics of mmWave signals directly into the learning process. We train a transformer-based shape completion network entirely on synthetic 3D point clouds (e.g., ShapeNet~\cite{shapenet}) while introducing a specularity-aware inductive bias that models the sparse, specular reflections of mmWave signals. Furthermore, our framework incorporates reflection-dependent visibility patterns, enabling the model to anticipate regions of the object that may be inherently unobserved.  Finally, to enhance robustness against real-world noisy measurements, we adapt the model's loss function to jointly refine the noisy input and complete missing surfaces.

We then incorporate this design into a real-world inference process that transforms raw mmWave signals into several candidate surface hypotheses, applies our physics-aware shape completion model to each, and performs entropy-guided surface selection to output a single high-fidelity 3D reconstruction. 

Our approach achieves state-of-the-art results on through-occlusion reconstruction. Across 61 diverse objects from the YCB dataset~\cite{ycb}, \name\ achieves a 72\% recall, an 18\% improvement over the next-best baseline, while maintaining a high precision of 85\%. 

\noindent\textbf{Contributions.}
We summarize our main contributions:
\begin{itemize}[itemsep=-5pt, topsep=0pt, leftmargin=*]
    \item \textbf{Physics-Aware mmWave Shape Completion.} We introduce the first through-occlusion mmWave 3D shape completion framework for diverse objects. It features a physics-aware training pipeline and real-world inference process, which together enable training entirely on synthetic data while showing 3D reconstruction on real-world data.

    \item \textbf{State-of-the-Art Performance.} On the real-world MITO dataset~\cite{mito}, we boost recall from 54\% to 72\% over existing mmWave reconstruction methods, (Backprojection~\cite{LauraEPFL}, mmNorm~\cite{mmnorm},  RMap~\cite{rmap}), while retaining 85\% precision. 

    \item \textbf{Ablation Study Against Vanilla Vision-Based Models.} We outperform vanilla vision-based completion models applied to mmWave partials, increasing recall by 12\% and achieving the highest precision of 85\%.

\end{itemize}

\vspace{-0.05in}
\section{Related Work}
\vspace{-0.075in}

\noindent
\textbf{Shape Completion in Visible Spectrum.}
Shape completion in the visible spectrum has been a long-standing problem in computer vision, with methods evolving from early geometric approaches~\cite{amenta1998new,amenta2001power,kazhdan2006poisson,alexa2001point} and encoder-decoder architectures~\cite{pcn,topnet} to more recent transformer- and diffusion-based frameworks~\cite{yu2021pointr,seedformer,snowflake,shapeformer,lyu2021conditional,du2025superpc,li2023proxyformer,kasten2023point,wei2025pcdreamer,bekci2025escape,zhou20213d}. These models achieve remarkable performance on synthetic, camera, and LiDAR datasets, benefiting from point clouds with relatively high coverage and low noise compared to mmWave partial observations. However, these models are not directly applicable to the mmWave domain, as they fail to capture the physical characteristics of mmWave propagation, such as specularity, low spatial resolution, and low signal-to-noise ratios (SNRs). Consequently, vision-based completion methods struggle to recover reliable geometry from mmWave point clouds (as we demonstrate empirically in Sec.~\ref{sec:results_ots}), underscoring the need for architectures that explicitly account for mmWave-specific physical properties.

\begin{figure*}[t]
\centering
    \includegraphics[width=\linewidth]{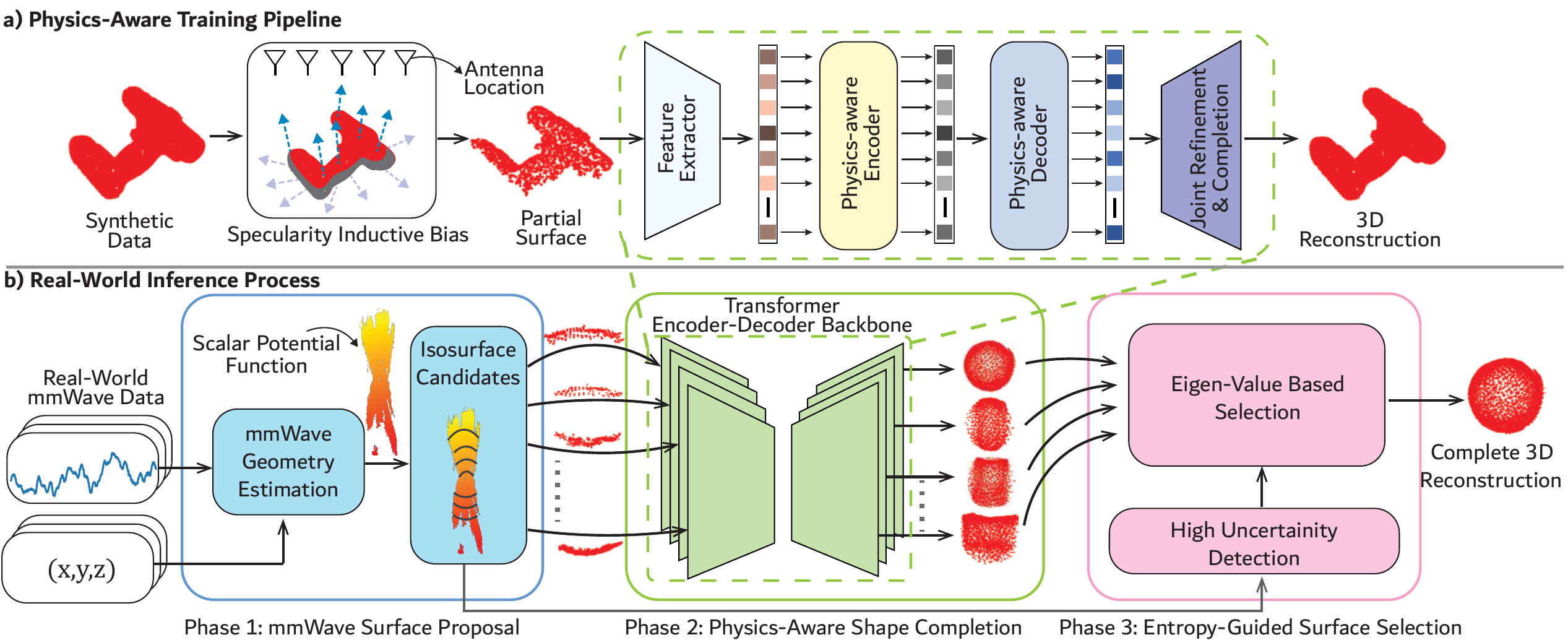} 
    \vspace{-0.275in}
    \caption{\footnotesize{\textbf{mmWave Reconstruction Pipeline.} \textnormal{a) \name's physics-aware training pipeline incorporates physical properties through a specularity-aware inductive bias, reflection-dependent visibility, and joint refinement and completion framework to enable training on entirely synthetic data. b) \name's real-world inference process leverages a three-stage pipeline to reconstruct a complete 3D object from real mmWave signals. }}}

    \vspace{-0.2in}
    \label{fig:tech}
\end{figure*}

\noindent
\textbf{mmWave Imaging.}
The vast majority of work in the mmWave space can image and reconstruct only the radar-facing surfaces of occluded objects or scenes, since they operate based on reflections that return from the surface. This includes volumetric~\cite{LauraEPFL,S1,laviada2017multiview,laviada2018multiview}, interferometric~\cite{SO7,SO8,SO9,SO10}, surface-normal based~\cite{mmnorm} estimation approaches, as well as learning-based image enhancement~\cite{Hawkeye,rostami2022deep,hrnet,mmid,mishape} and scene reconstruction~\cite{rmap,dreampcd,RadarHD,zhang2024towards,luan2024diffusion,c4rfnet}. This limitation is why airport mmWave scanners~\cite{airportscanner1,airportscanner2} require a large infrastructure to rotate antennas around the entire human body, and require a large non-commercially available bandwidth to achieve high-accuracy depth estimation. In contrast, \name\ is able to perform complete 3D reconstruction with limited coverage of an object and only commercially available bandwidth through a physics-aware shape completion pipeline. 

Additionally, some prior works have investigated mmWave full mesh reconstruction by focusing on a very limited set of objects (i.e., one to five categories) through strong data priors. For example, some works can only reconstruct humans due to their strong assumptions about human morphology~\cite{mmdear,mmmesh,xie2023mmpoint,yang2024mmbat,m4esh,immfusion,rf-avatar}. Similarly, other work can only reconstruct up to five categories of objects (chairs, cars, robot arms, boxes, and desks)~\cite{3drimr,r2p,sun20223d}. These works are fundamentally limited due to the fact that real-world and simulated mmWave data is extremely scarce. In contrast, \name\ eliminates the need for real-world mmWave training data by incorporating physical mmWave properties with readily-available  synthetic 3D datasets, thus enabling generalization to diverse objects.

\noindent
\textbf{Alternative Through-Occlusion Perception Modalities.}
Recovering geometry through occlusions has been explored with various sensing modalities. For example, X-rays can penetrate opaque materials but use ionizing radiation, making them unsuitable for extended human exposure~\cite{X-ray_risks}. Additionally, acoustic or ultrasound is impractical for imaging through closed boxes due to the change in material properties (i.e., impedance mismatch) between air and cardboard~\cite{kim2010sound}.\footnote{Gel minimizes the mismatch when imaging through human tissue~\cite{ultrasoundimaging}.} Furthermore, thermal imaging~\cite{he2023smart,ivavsic2019human,maeda2019thermal,ring2012infrared} can only image objects via heat, and thus is more suitable for imaging living things (due to their body temperature) rather than inanimate objects. Finally, around-the-corner laser~\cite{OP1,OP2,OP3,OP4,OP5} imaging can produce images in non-line-of-sight settings but cannot perceive through opaque obstacles, making them ill-suited for scenarios where objects are in a closed box or beneath clutter. In contrast, mmWave sensing combines material penetration and human safety, making it a practical modality for through-occlusion perception~\cite{mmnorm,he2024see,Hawkeye,LauraEPFL,m4esh,lai2024enabling,he2024see,lu2020see,wang2022wavesdropper}. 

\vspace{-0.1in}

\section{Preliminaries}
\vspace{-0.05in}

Here, we describe the process of classical mmWave imaging at a high-level, and refer readers to~\cite{mmwave_tutorial} for more details. 

\noindent \textbf{Forming Images from mmWave Signals.}
Millimeter-wave (mmWave) radars transmit a frequency-modulated continuous wave (FMCW) signal which travels through occluding materials and reflects off hidden objects before being received back at the radar. These reflections carry information about how far the wave traveled, as a frequency shift, and the direction from which they arrived, as phase shifts across sensor locations. Together, this allows us to reconstruct a 3D image of the radar-facing surface of the object, where the image value $S(v)$ at each voxel $v$ estimates the power reflected from that point in space~\cite{RFCapture}:

\vspace{-0.075in}
{\eqsize
\begin{equation} \label{eq:sar}
    S(v) = \sum_{k=1}^{N} \sum_{t=1}^{T} h_k(t) e^{j 2 \pi (2||p_k-v||)/\lambda_t}
\end{equation}
}
\vspace{-0.075in}

\noindent where $p_k$ is the k\tth\ sensor location,  
$\lambda_t$ is the  wavelength of the t\tth\ sample (out of $T$), and $h_k(t)$ is the t\tth\ sample of the time-domain baseband received signal from the k\tth\ location.

\noindent \textbf{Specularity.} One primary limitation of these 3D mmWave images is that they only measure the radar-facing surface of an occluded object. Specifically, since mmWave signals experience primarily specular (mirror-like) reflections~\cite{SpecularReflection}, many parts of an objects surface will reflect signals away from the sensor and thus are not present in the image. \name\ overcomes this limitation by bridging raw mmWave signals with physics-aware shape-completion. 

\vspace{-0.075in}
\section{Method}
\vspace{-0.075in}

We design \name\ to perform high-fidelity, complete 3D reconstruction of hidden objects leveraging mmWave signals. Our core insight lies in a physics-aware training pipeline (Fig.~\ref{fig:tech}a) to enable learning shape completion for mmWave properties using entirely synthetic data (Sec.~\ref{sec:training_pipeline}). We additionally propose a real-world inference process (Fig.~\ref{fig:tech}b) that (1) converts raw, mmWave signals into candidate partial surface representations (Sec.~\ref{sec:phase_1}), (2) performs physics-aware shape completion (Sec.~\ref{sec:phase_2}), and (3) identifies the optimal reconstruction (Sec.~\ref{sec:phase_3}). 

\vspace{-0.05in}
\subsection{Problem Definition}
\vspace{-0.05in}

Our goal is to reconstruct a complete 3D point cloud $\hat{F}$ of a fully occluded object from a sequence of raw, complex-valued, time-domain mmWave measurements. Let $H \in \mathbb{C}^{N \times T}$ denote the collection of all measurements acquired from $N$ known sensor positions $P \in \mathbb{R}^{N \times 3}$, where each sensor position yields $T$ temporal samples. Our pipeline learns a mapping $f_\theta : (H, P) \mapsto \hat{F}$.

\vspace{-0.1in}
\subsection{Physics-Aware Training Pipeline}\label{sec:training_pipeline}
\vspace{-0.05in}

Training a shape-completion model for mmWave signals is fundamentally different from training one for visible-light sensors, such as cameras or LiDAR. Millimeter-wave returns are specular, anisotropic, and noisy, while existing completion models are trained on vision-like partials, which implicitly assume diffuse reflections and wide, uniform coverage, making them ill-suited for mmWave inputs.

To overcome this, our key insight is to embed mmWave physics directly into the training data and objective, creating a physics-consistent observation model that teaches the network physically plausible mmWave shape completion. By doing so, we enable training entirely on readily-available synthetic 3D datasets while achieving state-of-the-art generalization to real mmWave signals, as shown in Sec.~\ref{sec:results_baselines}.

\vspace{-0.05in}
\subsubsection{Specularity-Aware Inductive Bias} \label{sec:inductive_bias}
\vspace{-0.05in}

Existing vision-based completion models inherently encode inductive bias consistent with visible light and incompatible with mmWave signals. This arises from their camera-like partial observations which assume diffuse reflections and wide coverage. To address this, we reformulate the inductive bias with physically consistent partials which emulate the specular reflections of mmWave signals. Specifically, for a given 3D model, we only include points in our partial which 1) produce a specular return and 2) lie on the radar facing surface (and are not further occluded by the object itself). Formally, we create partial observations $O$ as:

\begin{figure}[t]
\centering
    \includegraphics[width=\linewidth]{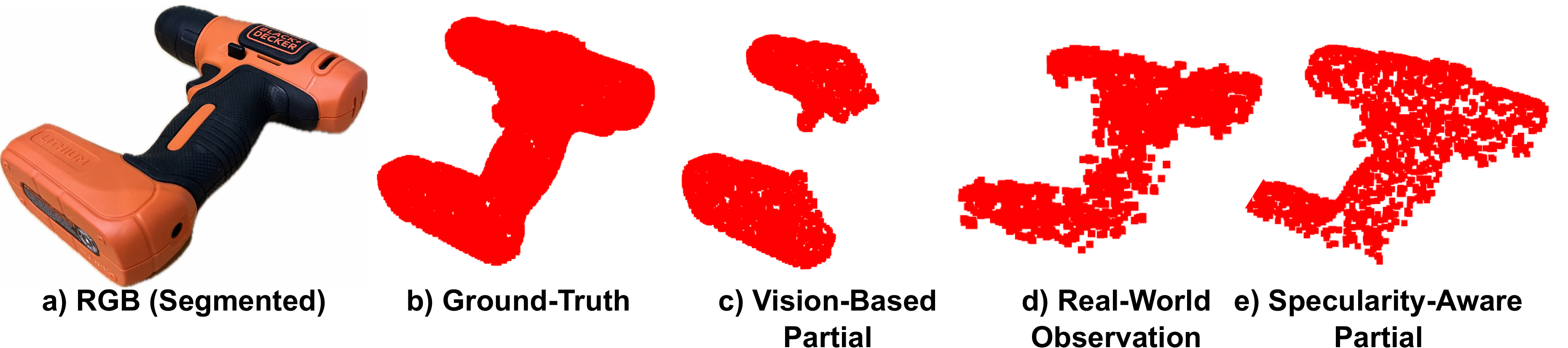} 
    \vspace{-0.25in}
    \caption{\footnotesize{\textbf{Specularity-Aware Inductive Bias.} \textnormal{Specularity-aware inductive bias generates training partials (e) that resemble real mmWave visibility (d), unlike standard masking used in vision-based models (c).}}}
    \vspace{-0.2in}
    \label{fig:specularity_aware}
\end{figure}

\vspace{-0.1in}
{\eqsize
\begin{equation}\label{eq:specularity}
    O=\bigg\{ s_i \in F \ \bigg| \theta_P(s_i) <\tau  \cap V(s_i)=1 \bigg\} 
\end{equation}}

\noindent where $s_i$ is a point in the full 3D cloud $F$. $\theta_P(s_i)$ denotes the smallest angular mismatch between $s_i$ and all sensor positions in $P$. It is defined as $\theta_P(s_i)=\min_{k\in P}\left|\acos\left( n_i\cdot u_{k,i} \right)\right|$ where $n_i$ is the normal at point $s_i$ and $u_{k,i}= (p_k-s_i)/||p_k-s_i||$ is the unit vector pointing from $s_i$ to $p_k$. $V(s_i)$ is 1 when a point $s_i$ lies on a radar facing surface and 0 otherwise.~\footnote{We compute this using Open3D's \texttt{hidden\_point\_removal}.}

Applying this inductive bias during training teaches the model to expect mmWave-specific partial observations, guiding it to focus on physically plausible surfaces. This results in more accurate reconstructions and better generalization to real-world mmWave measurements (See Sec.~\ref{sec:results_ablation}).

\vspace{-0.05in}
\subsubsection{Reflection-Dependent Visibility} \label{sec:augmentation}
\vspace{-0.05in}

Unlike optical sensors, mmWave visibility is strongly anisotropic: the measurable reflections depend on the incident angles and the reflection strength of an object. As a result, two objects with the same geometry may have markedly different visibility depending on their material properties. To model this behavior, we introduce a reflection-dependent visibility pattern that attenuates surface points according to physically-guided angular and material constraints. This replaces the common assumption of isotropic coverage and teaches the network that mmWave visibility is inherently non-uniform and angle dependent. Formally, for each specularity-aware partial $O$, we generate anisotropic partials $O_A$ as:

\vspace{-0.15in}
{\eqsize
\begin{align}\label{eq:fov}
% \vspace{-0.2in}
    O_{A} = \big\{s_i  \in & O \big| \theta_H(n_i)  < \tau_H 
    \ \cap 
   \theta_V(n_i) < \tau_V \big\}
% \vspace{-0.1in}
\end{align}}

\vspace{-0.075in}
\noindent where $\theta_H= \left|\acos\left(n_i\cdot [1,0,0]\right) \right|$ and $\theta_V= \left|\acos\left(n_i\cdot [0,1,0]\right) \right|$ are the angles of the specular return from point $s_i$ in the horizontal and vertical dimensions. $\tau_H$ and $\tau_V$ are parameters that can be changed to simulate different material properties. We train across a range of parameters to ensure robust mmWave completion.

Combined with our specularity-aware inductive bias, this visibility pattern further constrains the model to learn from physically plausible partial observations.

\vspace{-0.05in}
\subsubsection{Joint Refinement and Completion} \label{sec:jrac}
\vspace{-0.05in}

Existing vision-based shape completion models are designed for noise and resolution properties typical of a camera or LiDAR sensor, and therefore assume input partials can be directly concatenated with the reconstructed points. However, mmWave signals experience significantly higher noise levels  (e.g., up to 5$\times$ higher than depth-camera point clouds~\cite{mmnorm, kinect-acc}) and a reduction in resolution. Thus, existing concatenation strategies would propagate significant distortions into the final reconstruction. 

To address this, we introduce a joint refinement and completion method. Instead of preserving a noisy partial, we allow the model to simultaneously denoise and complete the object. To enable this, we incorporate noise during training to reflect the behavior of real-world mmWave signals. Then, we reformulate the loss function to allow the model to output the complete 3D shape (without concatenating the input), allowing it to reinterpret unreliable points rather than preserve them. Formally, we define the loss as the bi-directional chamfer distance between the complete, denoised output, $\hat{F}$ and the ground-truth shape, $F$:

{\eqsize
% \vspace{-0.15in}
\begin{equation}\label{eq:loss}
\vspace{-0.3in}
    \mathcal{L} = \frac{1}{|\hat{F}|} \sum_{s_i \in \hat{F}} \min_{g \in F} ||s_i - g|| + \frac{1}{|F|}\sum_{g \in F} \min_{s_i \in \hat{F}} ||g-s_i||
\vspace{0.2in}
\end{equation}}

This method further improves \name's generalization to real-world measurements, as we show emperically in Sec.~\ref{sec:results_ablation}

Together, these three techniques establish a new physics-aware learning paradigm for mmWave shape completion, enabling state-of-the-art through-occlusion reconstruction capabilities.

\noindent \textbf{Physics-Aware Encoder-Decoder}
We integrate the above techniques into a transformer-based encoder-decoder for shape completion which maps a partial observation to an estimate of the complete 3D geometry: $\hat{F}=f_p(O)$. Built on the PoinTr backbone~\cite{yu2021pointr}, our encoder–decoder is trained with a physics-consistent observation model (Eqs.~\ref{eq:specularity}~and~\ref{eq:fov}) and a denoising–completion objective (Eq.~\ref{eq:loss}), biasing the learned representation toward physically plausible mmWave surfaces. Our techniques enable us to train entirely with large, readily available synthetic 3D datasets~\cite{omniobject, toys4k, objaverse}, without relying on scarcely available  real-world mmWave data. Critically, this allows \name\ to generalize to many different objects.

\vspace{-0.05in}
\subsection{Real-World Inference Process}
\vspace{-0.05in}

After training a physics-aware shape completion model, we next develop an inference process that converts raw, real-world mmWave signals into complete 3D reconstructions (Fig.~\ref{fig:tech}b). Our pipeline follows three phases: mmWave surface proposal, physics aware shape completion, and entropy-guided surface selection.

\vspace{-0.05in}
\subsubsection{Phase 1: mmWave Surface Proposal} \label{sec:phase_1}
\vspace{-0.05in}

We first transform raw mmWave measurements into a set of candidate partial surfaces which accurately capture the geometric information contained in the reflections.

Typically, partial mmWave point cloud estimation relies on thresholding a mmWave 3D power image (Eq.~\ref{eq:sar}). However, this results in point clouds with significant erroneous points, as shown by the Backprojection baseline in Sec.~\ref{sec:results_baselines}. Instead, we leverage recent advancements in mmWave imaging~\cite{mmnorm} which can transform raw reflections into a space of geometrically consistent partial surfaces. 

Formally, we compute a scalar potential function $f(v)$ consistent with mmWave reflections, where each isosurface~\footnote{An isosurface is a surface of a constant value within a function.} of this function is one possible partial surface~\cite{mmnorm}:

% \vspace{-0.05in}
{\eqsize
\begin{equation}
\label{eq:scalar_field}
    f(v) = \sum_{j\in R} N(v_j) \cdot d_j
\end{equation}}

\noindent where $N(v_j)$ is the estimated mmWave normal vector field, and $f(v_0)=0$ at a reference voxel $v_0$, and $R$ is the discrete path connecting $v_0$ and $v$. The value $f(v)$ represents the accumulated field integral along $R$, and $d_j$ is the direction vector along $R$ pointing from $v_{j-1}$ to $v_j$

Unlike prior approaches which directly attempt to estimate the best candidate surface, we instead summarize the entire physically-plausible partial space by sampling candidate surfaces throughout this scalar function. This preserves available geometric information used during shape completion, and prevents critical errors caused by prematurely selecting the wrong surface. Formally:

\vspace{-0.05in}
{\eqsize
\begin{equation}
    C_{p,i} =\{ v\  |\  |f(v) - I(i)| <\delta \} \vspace{-0.01in}
\end{equation}}

\noindent where $C_{p,i}$ is the i\tth\ candidate partial surface, $I(i)$ is the value of the i\tth\ isosurface we sample, and $\delta$ is a small constant to account for numerical discrepancies. 

\vspace{-0.05in}
\subsubsection{Phase 2: Physics-Aware Shape Completion}\label{sec:phase_2}
\vspace{-0.05in}

Next, we convert the space of partial observations into a space of complete 3D reconstructions. We apply our physics-aware shape completion model, trained to incorporate physical properties as described in Sec.~\ref{sec:training_pipeline}, to each candidate partial surface. This produces a set of physically-plausible candidate complete reconstructions, $C_{F,i}$:

{\eqsize
\begin{equation}
    % \vspace{-0.05in}
    C_{F,i} = f_p(C_{p,i})
    % \vspace{-0.05in}
\end{equation}}

\newcolumntype{C}[1]{>{\centering\arraybackslash}p{#1}}

\vspace{-0.05in}
\subsubsection{Phase 3: Entropy-Guided Surface Selection}
\label{sec:phase_3}
\vspace{-0.05in}

In our final phase, we identify the optimal reconstruction from the space of candidate reconstructions.

Under high signal-to-noise conditions, we follow prior work~\cite{mmnorm} and select the optimal surface by comparing the simulated mmWave response of each candidate to the measured signals. 
However, when reflections are weak and the normal field contains significant noise, this often leads to erroneous surface selection. To handle these challenging cases, we introduce an entropy-guided selection strategy.

\noindent\textbf{Detecting High-Uncertainty Reconstructions.}
We identify partial observations corrupted by noise, where weak reflections or noisy normals cause the scalar field $f(v)$ (Eq.~\ref{eq:scalar_field}) to produce irregular, vertically stacked voxels instead of smooth isosurfaces. We use the vertical spread of these voxels to serve as a proxy for reconstruction uncertainty. Formally, we classify reconstructions as high uncertainty when $\frac{|C_{p,i}^{xy}|}{|C_{p,i}|}>0.6$, where $C_{p,i}^{xy}$ is constructed by projecting the observation $C_{p,i}$ to 2D and removing duplicate points.

\noindent \textbf{Determining the Optimal Reconstruction. }
For high-uncertainty cases, we observe that physically consistent inputs produce  continuous, locally planar reconstructions, whereas inaccurate or noisy inputs yield high-entropy point clouds dispersed over a larger spatial volume. We therefore select the candidate reconstruction with the lowest degree of local entropy.

We quantify entropy by sampling local neighborhoods~\footnote{We use kNN with k=30.} and computing the covariance eigenvalues $\lambda_1\geq \lambda_2 \geq \lambda_3$. Since planar regions have two dominant eigenvalues, whereas high-entropy regions have three of similar magnitude, we define an entropy score $\lambda_3/\lambda_1$~\cite{eigenvalues}. We then combine this score across all neighborhoods, and select the reconstruction with the smallest ratio:

\vspace{-0.1in}
{\eqsize
\begin{equation}
i^{\star} = \arg\min_i \left(
\texttt{pcntl}\left(\left\{\frac{\lambda_{3,p}}{\lambda_{1,p}} \forall p \in C_{F,i} \right\}, 75\right) \right)
\label{eq:entropy_selection}
\end{equation}}
% \vspace{-0.05in}

Then, our final surface selection is $\hat{F}=C_{F,i^{\star}}$. 
% \vspace{0.05in}

\noindent Together, these three phases together output a single, high-fidelity 3D reconstruction of a fully occluded object.

\begin{figure*}[t]
\centering
    \includegraphics[width=\linewidth]{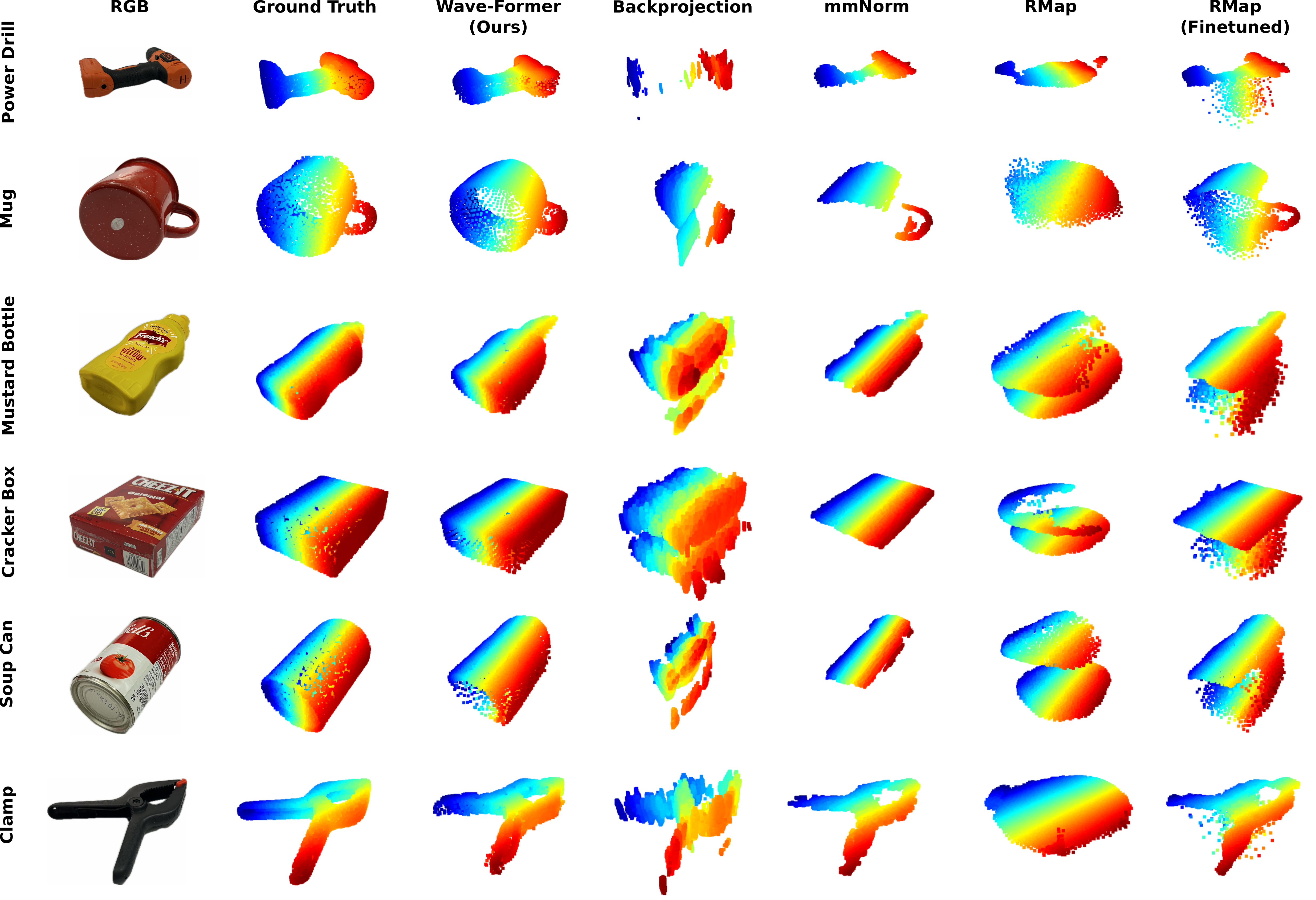} 
    \vspace{-0.3in}
    \caption{\footnotesize{\textbf{Qualitative Results.} \textnormal{Visual comparison of mmWave 3D reconstruction on real-world, fully occluded objects. State-of-the-art baselines suffer from artifacts such as high noise and limited coverage, while \name\ consistently reconstructs shapes with high fidelity. }}}

    \vspace{-0.15in}
    \label{fig:result_qualitative}
\end{figure*}

\vspace{-0.1in}
\section{Experiments} \label{sec:results}
\vspace{-0.05in}

\subsection{Dataset and Implementation}
\vspace{-0.05in}
\noindent \textbf{Datasets:} To train \name, we leverage three publicly available 3D object datasets: OmniObject3D~\cite{omniobject}, Toys4K-3D~\cite{toys4k}, and the Thingiverse subset of Objaverse~\cite{objaverse}
, totaling over 25K 3D object point clouds.

We evaluate \name\ on a real-world dataset of mmWave measurements, MITO~\cite{mito}. The dataset contains measurements of 61 objects from the YCB dataset~\cite{ycb}. These items consist of a diversity of tasks (kitchen items, tools, food items, toys, etc), materials (wood, metal, cardboard, plastic, etc), and geometries (complex shapes, sharp edges, flat and curved surfaces, etc). 
This includes experiments in both line-of-sight and fully occluded settings for each object.
More details can be found in~\cite{mito}.

\begin{figure*}
\small
\setlength{\tabcolsep}{1mm}
\begin{minipage}[t]{0.38\linewidth}
\centering
\begin{tabular}{l | C{.9cm} C{.9cm} C{1.5cm} C{1.1cm}}
\toprule
& CD $\downarrow$ & FS   $\uparrow$ & Precision $\uparrow$ & Recall $\uparrow$ \\
\midrule
Backprojection~\cite{LauraEPFL}  &    0.180 &    40\% &    43\%  &    45\%  \\
mmNorm~\cite{mmnorm}         &    0.214 &    45\% &    \textbf{89\%}  &    34\%\\
R-Map~\cite{rmap}           &    0.273 &    23\% &    40\%  &    17\%  \\
R-Map (Finetuned)~\cite{rmap}     &    0.330 &    62\% &    81\%  &    54\%  \\
\midrule
\textbf{\name}                    &  \textbf{0.069} & \textbf{75\%} & 85\% & \textbf{72\%} \\
\bottomrule
\end{tabular}
\vspace{-0.1in}
\captionof{table}{\footnotesize{Comparison between \name\ and state-of-the-art mmWave reconstruction baselines. }}
\vspace{-0.25in}
\label{tab:baselines}
\end{minipage}
\hfill
\begin{minipage}[t]{0.5\linewidth}
\centering
% \begin{tabular}[t]{l | c c c c}
\begin{tabular}{l | C{.9cm} C{.9cm} C{1.5cm} C{1.1cm}}
% \small
\toprule
& CD $\downarrow$ & FS $\uparrow$& Precision $\uparrow$ & Recall $\uparrow$\\
\midrule
mmNorm + PoinTr             &    0.104 &    62\% &    81\%  &    53\% \\
mmNorm + SnowFlakeNet      &    0.097 &    66\% &    80\%  &    60\%  \\
mmNorm + SeedFormer        &    0.095 &    66\% &    83\%  &    59\% \\
mmNorm + PCN               &    0.138 &    58\% &    70\%  &    56\% \\
\midrule
\textbf{\name}                     &  \textbf{0.069} & \textbf{75\%} & \textbf{85\%} & \textbf{72\%} \\
\bottomrule
\end{tabular}
\vspace{-0.1in}
\captionof{table}{\footnotesize{Comparison with state-of-the-art vision-based vanilla shape completion models applied to the state-of-the-art mmWave reconstruction method.}}
\vspace{-0.25in}
\label{tab:ots}
\end{minipage}
\end{figure*}

\noindent \textbf{Training:} We train our model using exclusively synthetic data, using the techniques in Sec~\ref{sec:training_pipeline}. Each of the synthetic datasets are shuffled and split into 80\% for training and 20\% for testing. Additional details can be found in A.1. 

\noindent \textbf{Evaluation:} For real-world evaluation, we directly input the raw mmWave radar data into \name\ without any additional masking or augmentations. We compare to the ground-truth 3D reconstructions provided in the MITO dataset~\cite{mito}. 
Following prior work~\cite{yu2021pointr,snowflake}, we align the reconstructed partial point clouds generated in Sec.~\ref{sec:phase_1} with the ground-truth point cloud for inference and evaluation.

\noindent \textbf{Evaluation Metrics:} We adopt four standard metrics~\cite{3d_fscore}: Chamfer Distance (CD), Precision, Recall, and F-Score, which jointly assess the accuracy and completeness of the reconstructions. We provide detailed definitions in A.2.

\noindent \textbf{Baselines:} We evaluate our approach against four state-of-the-art mmWave reconstruction baselines. We include detailed descriptions in A.3:

% \vspace{-0.075in}
\begin{itemize}[itemsep=-5pt, topsep=0pt, leftmargin=*]
    \item \textit{Backprojection~\cite{LauraEPFL}:} This is the classical and most widely used mmWave imaging approach. It is a first-principle method for volumetric mmWave reconstruction.
     \item \textit{mmNorm~\cite{mmnorm}:}  This is a recently-introduced state-of-the-art mmWave 3D reconstruction approach. It is also a first-principle method which estimates surface normal vectors and uses them to reconstruct the surface of the object
     \item \textit{RMap~\cite{rmap}:} This is a state-of-the-art learning-based method for mmWave reconstruction, originally designed for scene level understanding. This model expects mmWave point cloud inputs, so we use mmNorm to create the input. % to this model.
     
     \item \textit{RMap (Finetuned)~\cite{rmap}:} We finetune RMap for object reconstruction on the same training data as \name.
\end{itemize}
% \vspace{-0.05in}

\vspace{-0.075in}
\subsection{Baseline Comparisons} \label{sec:results_baselines}
\vspace{-0.05in}

\subsubsection{Qualitative Results}
\vspace{-0.025in}
To start, we compare qualitative results across \name\ and all four baselines using real-world measurements. Fig.~\ref{fig:result_qualitative} shows an isometric view of the ground-truth RGB (segmented) and point cloud, as well as the reconstruction for each method across several fully-occluded objects.
\name\ is able to consistently reconstruct the complete shape of the object, even for complex geometries such as a power drill and clamp. In contrast, the baselines suffer from low accuracy, limited coverage, high noise, and, in some cases, even a complete a lack of discernible geometry. 
These results demonstrate the significant advancement of \name\ over prior state-of-the-art 
mmWave 3D reconstruction methods.

\vspace{-0.05in}
\subsubsection{Quantitative Results}
\vspace{-0.05in}
Table~\ref{tab:baselines} reports the average chamfer distance, F-Score, precision, and recall across \name\ and all baselines. Notably, \name\ achieves a significant increase in recall, from 54\% to 72\% over the best-performing baseline, RMap (Finetuned), while maintaining a high precision of 85\%.~\footnote{\textred{mmNorm shows slightly higher precision since, as a first-principles method, it does not infer full geometry, leading to much lower recall.}\cut{It is important to note that mmNorm achieves a slightly higher precision given it is a first principle based method that does not attempt to infer the entire object geometry (as evident by its significantly worse recall).}} Additionally, \name\ achieves the lowest chamfer distance of 0.069, compared to 0.18 for the best-performing baseline. This demonstrates the value of our techniques for enabling complete and high-accuracy 3D reconstruction of fully occluded objects.

\vspace{-0.08in}
\subsection{Comparison to Vision-Based Shape Completion}\label{sec:results_ots}
\vspace{-0.05in}

We also evaluate whether state-of-the-art \textred{vanilla} vision-based shape completion models can achieve high-accuracy mmWave 3D reconstruction. \textred{To produce radar-visible partial point clouds, we use the state-of-the-art mmWave reconstruction baseline with the highest precision per Table~\ref{tab:baselines} (mmNorm), which is then fed into vision-based completion models.} To ensure a fair comparison, all models are fine-tuned with the same synthetic datasets as used to train \name, while applying their own partial point cloud creation strategies. Table~\ref{tab:ots} reports the performance of \name\ compared to four state-of-the-art models. \name\ outperforms all models across all metrics, achieving an increase in recall from 60\% to 72\%, while also achieving the highest precision of 85\%. This shows the importance of incorporating  physical properties into the shape completion model. \textred{We also provide qualitative examples of \name\ compared to each vanilla shape completion model in A.4.}

\vspace{-0.08in}
\subsection{Ablation Study}\label{sec:results_ablation}
\vspace{-0.05in}

Next, we analyze how different components of \name's design contribute to the overall performance. Table~\ref{tab:ablation} shows the average and 75\tth\ percentile CD, as well as the marginal percent increase,  of \name\ compared to three different partial implementations. First, when removing our specularity-aware inductive bias and reflection-dependent visibility (model A), we see a significant drop in performance, with the average chamfer distance increasing by 52\% and the 75\tth\ percentile increasing by 67\%.  Next, when also removing our joint reconstruction and completion (model B), average chamfer distance increases by an additional 10\%. Finally, when also removing our entropy-aware surface selection (model C), the 75\tth\ percentile continues to increase an additional 19\%. \textred{We further demonstrate the benefit of our entropy-aware surface selection through qualitative examples in A.5, which show that for some objects, our selection process leads to drastically improved reconstructions.} Together, these results demonstrate how each component of \name\ contributes to its overall performance.

\begin{table}[h]
\small 
\setlength{\tabcolsep}{0.55mm}
\vspace{-0.15in}
\begin{tabular}{l| c c c | c c | c c}
\toprule
 &  &  &  & \multicolumn{2}{c|}{Average} &  \multicolumn{2}{c}{75\tth\ Pcntl}  \\
Model & Physics & JRAC & Entropy & CD & \% Inc. & CD & \% Inc.\\
\midrule
{\footnotesize \name} & \checkmark& \checkmark& \checkmark &\textbf{0.069} & -- & \textbf{0.072} & -- \\
A  & & \checkmark & \checkmark &     0.105 &  \textcolor{ForestGreen}{52\%} &  0.120 & \textcolor{ForestGreen}{67\%} \\
B    & &&\checkmark    & 0.115 & \textcolor{ForestGreen}{10\%} &   0.122 & \textcolor{ForestGreen}{2\%} \\
 C    &&&    & 0.116 & \textcolor{ForestGreen}{1\%} &   0.145 & \textcolor{ForestGreen}{19\%} \\
\bottomrule
\end{tabular}
\vspace{-0.1in}
\caption{\footnotesize{Ablation study of average and 75\tth\ percentile CD, as well as marginal \% increase, for different components of \name.}}
\vspace{-0.275in}
\label{tab:ablation}
\end{table}

\subsection{Microbenchmarks}
We performed microbenchmark experiments to understand the
impact of various factors on \name.

\vspace{0.055in}
\subsubsection{Impact of Occlusions}\label{sec:nlos}

\begin{table}[h]
\small 
\setlength{\tabcolsep}{0.5mm}
\vspace{-0.17in}
\begin{tabular}{l | c c | c c }
\toprule
& \multicolumn{2}{c|}{Line-of-Sight} & \multicolumn{2}{c}{Fully Occluded}\\
& \ \ \ CD $\downarrow$ \ \ \ & F-Score $\uparrow$ & \ \ \ CD $\downarrow$\ \ \ & F-Score $\uparrow$\\
\midrule
Backprojection~\cite{LauraEPFL} & 0.177 & 40\% & 0.183 &40\%  \\
mmNorm~\cite{mmnorm}          & 0.198 & 46\% &0.231 &44\%  \\
R-Map~\cite{rmap}            & 0.273    & 23\%   & 0.272    & 24\%  \\
R-Map (Finetuned)~\cite{rmap}          & 0.091 & 64\% & 0.581 & 61\%  \\

\midrule
\name                     & \textbf{0.058} & \textbf{78\%}& \textbf{0.080} &\textbf{73\%} \\
\bottomrule
\end{tabular}
\vspace{-0.1in}
\caption{\footnotesize{Average performance in line-of-sight \& fully occluded settings.}}
\vspace{-0.17in}
\label{tab:nlos}
\end{table}

First, we compare the performance of \name\ and all baselines in line-of-sight and fully-occluded settings. Table~\ref{tab:nlos} reports the average chamfer distance and F-Score. 
Notably, \name\ performs similarly in both settings, experiencing only a minor decrease in performance in fully-occluded settings.
This demonstrates the ability of \name\ to not only reconstruct objects in visual line of sight, but also objects that are fully hidden from view.

\vspace{-0.03in}
\subsubsection{Impact of Object Size}
% \vspace{-0.05in}

\begin{table}[h]
\small 
\setlength{\tabcolsep}{3.2mm}
\vspace{-0.125in}
\begin{tabular}{l | c | c | c  }
\toprule
& Large & Medium & Small\\
\midrule
Backprojection~\cite{LauraEPFL}& 0.092  & 0.152  &  0.293  \\
mmNorm~\cite{mmnorm}          & 0.078  & 0.201  &  0.361 \\
R-Map~\cite{rmap}           & 0.136  & 0.239   & 0.452  \\
R-Map (Finetuned)~\cite{rmap}           & 0.057 & 0.621 & 0.158 \\
\midrule
\name                    & \textbf{0.044} & \textbf{0.057} & \textbf{0.109}  \\
Improvement &  \textcolor{ForestGreen}{0.013} & \textcolor{ForestGreen}{0.095}  & \textcolor{ForestGreen}{0.049} \\
\bottomrule
\end{tabular}
\vspace{-0.1in}
\caption{\footnotesize{Average CD for different object sizes.}}\label{tab:size}
\vspace{-0.17in}
\end{table}

Next, we evaluate the performance of \name\ and all baselines across objects of different sizes.
We divide all objects into three categories based on their longest dimension: Large ($>$20~cm), Medium (10~cm-20~cm), and Small ($<$10~cm). Table~\ref{tab:size} reports the average chamfer distance across different object sizes. As expected, \name\ performs best for large objects, with a chamfer distance of 0.04, compared to 0.06 for the best-performing baseline. 
As objects decrease in size, mmWave perception becomes increasingly challenging (See A.6).

\vspace{-0.03in}
\subsubsection{Impact of mmWave Coverage}

% \vspace{-0.1in}

\begin{table}[h]
\small 
\vspace{-0.1in}
\setlength{\tabcolsep}{1.8mm}
\begin{tabular}{l | c | c | c  }
\toprule
& Moderate & Challenging & Extreme\\
\midrule
Backprojection~\cite{LauraEPFL} & 0.076  & 0.140 & 0.325\\
mmNorm~\cite{mmnorm}           & 0.064 & 0.155  & 0.434 \\
R-Map~\cite{rmap}               & 0.126 & 0.221    & 0.481 \\
R-Map (Finetuned)~\cite{rmap}           & 0.050 & 0.076  & 0.904 \\
\midrule
\name                    & \textbf{0.033} &\textbf{0.048}& \textbf{0.126}\\
Improvement & \textcolor{ForestGreen}{0.017} & \textcolor{ForestGreen}{0.028}  & \textcolor{ForestGreen}{0.199}\\
\bottomrule
\end{tabular}
\vspace{-0.1in}
\caption{\footnotesize{Average CD for different initial object coverage.
}}
\vspace{-0.1in}
\label{tab:difficulty}
\end{table}

Finally, we assess how varying levels of coverage affect the performance of \name. 
To start, we compute the percent of ground-truth points “covered” by the mmWave partial observation, where a point is considered covered if its nearest neighbor in the input lies within 0.08~m (for objects scaled to a unit sphere).
We then define three categories of objects: Moderate ($>$36\%), Challenging (18\%-36\%), and Extreme ($<$18\%).~\footnote{Thresholds were selected for balanced category distribution.} It is important to note that these percentages are significantly more challenging than most standard vision-based shape completion models, which only consider point clouds with 25\% to 75\% coverage~\cite{yu2021pointr,seedformer}. This is due to the specular reflections of mmWave signals, which result in significantly lower coverage than visible light.

Table~\ref{tab:difficulty} reports the average chamfer distance of \name\ and all baselines. In all categories, \name\ outperforms all baselines. As the objects become more challenging, \name\ outperforms the best baseline by an increasing amount, with a 0.2 improvement for the Extreme category. This highlights the benefit of our techniques for enabling through-occlusion 3D mmWave reconstruction even on extremely challenging examples. \textred{We also show qualitative examples from each category in A.7.}

\vspace{-0.07in}
\section{Conclusion \& Future Opportunities}
\vspace{-0.07in}

We presented \name, the first mmWave 3D reconstruction method that can operate on fully-occluded, diverse, everyday objects. \name\ leverages a novel physics-aware training pipeline to integrate core physical properties of mmWave signals into the completion process. Then, \name\ uses a real-world inference process which proposes candidate surfaces, leverages our physics-aware shape completion model, and selects a final reconstruction. These techniques allow \name\ to be trained on entirely synthetic 3D data while demonstrating impressive generalization to real-world signals. 

Building on these contributions, we believe \name\ opens up many exciting directions for future research. For example, it would be interesting to use \name\ to develop novel downstream capabilities, such as fully-occluded robotic grasping, augmented reality with through-occlusion perception, or automatic package verification in shipping and logistics. Additionally, as the first paper to tackle the challenge of diverse mmWave 3D reconstruction, we hope that \name\ inspires future research \textred{towards reconstruction}\cut{which can reconstruct shapes} with even higher fidelity or using increasingly smaller radar scans. Finally, \textred{our techniques for training entirely on synthetic data may bring benefit to other mmWave tasks, such as fully-occluded classification, segmentation, and pose estimation. }\cut{we imagine our techniques which enable training on entirely synthetic data can be used to benefit other mmWave-driven tasks, such as fully-occluded classification, segmentation, and 6DoF pose estimation. }

{\looseness=-1
More generally, we hope this work marks an important step towards generalizable through-occlusion perception. }

{
    \small
    \bibliographystyle{ieeenat_fullname}
    \bibliography{ourbib}
}

\end{document}